\pdfoutput=1

\documentclass[11pt]{article}

\usepackage{acl}



\usepackage{amssymb}
\usepackage{times}
\usepackage{latexsym}
\usepackage{listings}
\usepackage{graphicx}
\usepackage{subcaption}
\usepackage{xcolor}
\usepackage[utf8]{inputenc}
\usepackage{microtype}
\usepackage{inconsolata}
\usepackage{graphicx}
\usepackage{amsmath}
\usepackage{booktabs}
\usepackage{float}
\usepackage{tabularx} 
\usepackage{multirow}
\usepackage{booktabs}
\usepackage{xurl}

\usepackage{xspace} 

\newcommand{\qafsc}{Q\textsc{uite}AF\textsc{ew}\xspace}

\usepackage{tcolorbox}

\tcbuselibrary{breakable}

\definecolor{lightorange}{HTML}{FFE0B2}

\newcommand{\token}[1]{
  \colorbox{lightorange}{\texttt{#1}}
}

\newtcolorbox{textttbg}{
  breakable,             
  colback=lightorange,   
  colframe=lightorange,  
  boxrule=0pt,           
  arc=0mm,               
  left=2pt,              
  right=2pt,             
  top=2pt,               
  bottom=2pt,            
  boxsep=0pt,            
  fontupper=\ttfamily,   
  parbox=false,          
  before upper={\parindent0pt\noindent} 
}

\title{One Task Vector is not Enough: \\A Large-Scale Study for In-Context Learning}

\author{Pavel Tikhonov \\
  AIRI / Moscow, Russia \\
  Skoltech / Moscow, Russia \\
  \texttt{tikhonov@airi.net} \\\And
  Ivan Oseledets \\
  AIRI / Moscow, Russia \\
  Skoltech / Moscow, Russia \\
  \texttt{oseledets@airi.net} \\\And
  Elena Tutubalina \\
  AIRI / Moscow, Russia \\
  \texttt{tutubalina@airi.net} \\}

\lstdefinestyle{promptstyle}{
    basicstyle=\ttfamily\footnotesize, 
    frame=single,
    rulecolor=\color{black},
    breaklines=true,
    backgroundcolor=\color{gray!10},
    showstringspaces=false,
    tabsize=2,
    captionpos=b,
    aboveskip=1em,
    belowskip=1em
}

\begin{document}

\maketitle

\begin{abstract}
In-context learning (ICL) enables Large Language Models (LLMs) to adapt to new tasks using few examples, with task vectors—specific hidden state activations—hypothesized to encode task information. Existing studies are limited by small-scale benchmarks, restricting comprehensive analysis. We introduce \qafsc{}, a novel dataset of 3,096 diverse few-shot tasks, each with 30 input-output pairs derived from the Alpaca dataset. Experiments with Llama-3-8B on \qafsc{} reveal: (1) task vector performance peaks at an intermediate layer (e.g., 15th), (2) effectiveness varies significantly by task type, and (3) complex tasks rely on multiple, subtask-specific vectors rather than a single vector, suggesting distributed task knowledge representation.

\end{abstract}

\begin{table}[t]
\centering
\small
\begin{tabular}{@{} p{1.9cm} p{6cm} @{}}
\toprule
\textbf{Work} & \textbf{Task Categories \& Examples} \\
\midrule
\citet{hendel2023incontext} \newline (11 tasks) 
& \textit{Algorithmic}: Next letter, List first, List last, To uppercase\newline
  \textit{Translation}: Fr $\rightarrow$ En, Es $\rightarrow$ En\newline
  \textit{Linguistic}: Present $\rightarrow$ Gerund,  Singular $\rightarrow$ Plural\newline
  \textit{Knowledge}: Country $\rightarrow$ Capital, Person $\rightarrow$ Language \\

\midrule
\citet{kharlapenko2024interpreting} \newline (9 tasks)
& \textit{Linguistic}: Antonyms, Present Tense $\rightarrow$ Past Tense\newline
  \textit{Translation}: En $\rightarrow$ Es, En $\rightarrow$ Fr, Es $\rightarrow$ En\newline
  \textit{Knowledge}: Country $\rightarrow$ Capital, Person $\rightarrow$ Profession, Location $\rightarrow$ Language, Location $\rightarrow$ Religion \\

\midrule
\citet{luo2024task} \newline (6 tasks) 
& \textit{Knowledge}: Country $\rightarrow$ Capital, Country $\rightarrow$ Currency, Animal $\rightarrow$ Latin, Animal $\rightarrow$ Young, Food $\rightarrow$ Color, Food $\rightarrow$ Flavor \\

\midrule
\citet{todd2024function} \newline (Over 40 tasks) 
& \textit{Linguistic (e.g.)}: Antonyms, Present $\rightarrow$ Past, Singular $\rightarrow$ Plural\newline
  \textit{Knowledge (e.g.)}: Country $\rightarrow$ Capital\newline
  \textit{Translation (e.g.)}: English $\rightarrow$ French\newline
  \textit{Text Manipulation (e.g.)}: Capitalize \\
\midrule
\qafsc{} (Ours) \newline (3,096 tasks)
& Split into categories by the first word of the task:\newline
  \textit{Given}: 294 tasks \newline
  \textit{Generate}: 193 tasks \newline
  \textit{Rewrite}: 178 tasks \newline
  \textit{Create}: 159 tasks \newline
  \textit{Classify}: 125 tasks \newline
  \textit{Identify}: 110 tasks \newline
  \textit{Write}: 107 tasks \newline
  \textit{Find}: 99 tasks \newline
  \textit{Other}: 1,657 tasks \\
\bottomrule
\end{tabular}
\caption{Task Dataset Comparison on Task Vectors investigation.}
\label{tab:datasets_sizes_comparison}
\end{table}

\section{Introduction}

Transformer-based Large Language Models (LLMs) \citep{vaswani2017attention} excel at in-context learning (ICL), adapting to new tasks via a few prompt-based examples without weight updates \citep{brown2020language} and have shown impressive empirical results~\citep{liu2023pre, dong2022survey}. This capability enables rapid task adaptation; however, how LLMs internally represent and apply task information remains unclear. Recent work points to ``task vectors'' \citep{hendel2023incontext} or ``function vectors'' \citep{todd2024function} -- specific hidden state activations -- as the mechanism for encoding task rules.

\begin{table*}[!t]
\centering
\small
\begin{tabular}{@{} p{0.25\linewidth} p{0.25\linewidth} p{0.10\linewidth} p{0.3\linewidth} @{}}
\toprule
\textbf{Instruction} & \textbf{Example Input} & \textbf{Category} & \textbf{Explanation} \\
\midrule
Answer this question with a yes or no. & Will I be able to go to the park tomorrow? & INVALID & Requires future knowledge or personal context that an AI cannot predict \\
\addlinespace
Find a good restaurant near the given address & 660 Lexington Avenue, New York, NY 10022 & INVALID & Needs real-world data; "good restaurant" is subjective \\
\addlinespace
What is the largest city on this continent? & Africa & LIMITED & Limited size of a category, insufficient for 30+ diverse examples \\
\bottomrule
\end{tabular}
\caption{Examples of Alpaca entries filtered out due to being unsuitable for few-shot generation.}
\label{tab:alpaca_issues_examples}
\end{table*}

Prior studies, such as \citet{hendel2023incontext}, suggest that ICL compresses demonstration sets into task vectors that guide query processing. \citet{todd2024function} used causal analysis to locate these vectors, showing they capture semantic task aspects. While techniques like sparse autoencoders (SAEs) have begun to shed light on the interpretable features within a given task vector \citep{kharlapenko2024interpreting}, the fundamental question of whether a single such vector suffices for complex, multi-faceted tasks remains largely unexplored. \citet{luo2024task} extended this to vision-language models, demonstrating that task vectors are cross-modal, clustering by task rather than input modality (e.g., text or image) and emerging at intermediate layers to summarize tasks before generation. On some tasks, task vectors achieve near-excellent performance, often over 90\% accuracy. However, current studies mainly utilize toy, manually crafted datasets (see Tab.~\ref{tab:datasets_sizes_comparison}), which limits our understanding of task vector dynamics in diverse, large-scale settings.

To address this gap, we introduce \qafsc{}, a novel dataset comprising 3,096 diverse few-shot learning tasks, each with 30 unique input-output pairs derived from the Alpaca dataset \citep{alpaca}. This dataset spans a broad spectrum of tasks, from algorithmic operations to open-ended generative challenges, enabling a comprehensive exploration of in-context learning (ICL). Through experiments with \cite{grattafiori2024llama} on \qafsc{}, we uncover key insights into task vector dynamics. Task vector performance consistently peaks at a specific (such as the 15th on Llama-3-8B) intermediate layer, across diverse task categories like algorithmic processing and text rewriting. However, the effectiveness of single task vectors varies significantly depending on the task type, with some categories demonstrating robust results while others experience notable declines. Our analysis reveals that instead of relying on a single task vector, models utilize multiple subtask-specific vectors, indicating a more distributed task representation within the model.

\section{Methodology}
\label{sec:methodology}

\subsection{Introduction to Task Vectors}
\label{sec:task_vectors}
Formally, a task vector is the hidden state at a designated layer for a specific token in the few-shot prompt, often the separator token (e.g., \token{->}) marking the transition from input to output. For a prompt with $k$ input-output pairs (e.g., \texttt{big -> small}), the task vector $v_l$ is extracted as the hidden state at layer $l$ after processing the final token \token{->}.

To apply a task vector, we employ a causal intervention during zero-shot inference. For a new input (e.g., \texttt{hot ->}), the model processes the input up to the token \token{->}, at which point the hidden state at layer $l$ is replaced with $v_l$. The model then generates the output autoregressively, using this modified hidden state as part of its standard computation.

\subsection{Dataset Collection}
\label{sec:dataset_collection}

We built \qafsc{} by expanding the Alpaca dataset \citep{alpaca}, which contains instruction-following entries from OpenAI's text-davinci\_003. Many Alpaca entries include an \texttt{instruction} paired with an example \texttt{input} and \texttt{output}, providing a structure ideal for generating diverse few-shot learning tasks.
The \texttt{instruction} (e.g., ``Rewrite the given sentence to incorporate a hyperbole'') specifies the task, while the example \texttt{input} (e.g., ``The house was very old.'') and \texttt{output} (e.g., ``The house was older than the hills.'') demonstrate the expected transformation, enabling the creation of varied examples (e.g., ``The water was very cold.'' $\to$ ``The water was colder than the depths of Antarctica.''). 
The instructions cover a wide range of tasks, and could be categorized by their initial verb (e.g., ``generate'', ``rewrite'', ``classify''), as shown in Tab.~\ref{tab:datasets_sizes_comparison}.

\begin{figure}[t]
    \centering
    \includegraphics[width=0.48\textwidth]{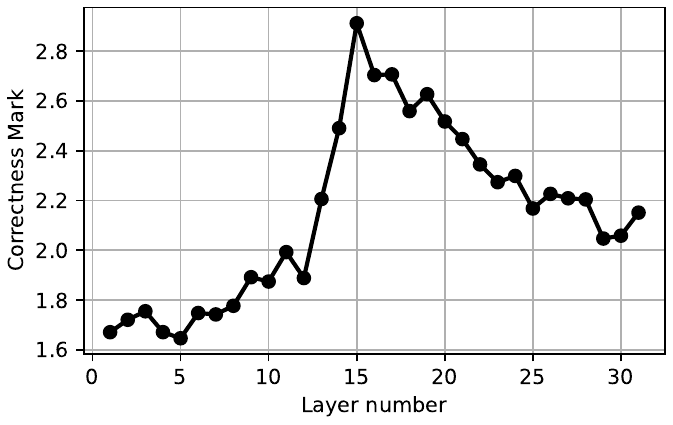}
    \caption{Average task vectors performance on \qafsc{} dataset.}
    \label{fig:overall_performance}
\end{figure}

\begin{figure*}[h]
    \centering
    \includegraphics[width=0.9\textwidth]{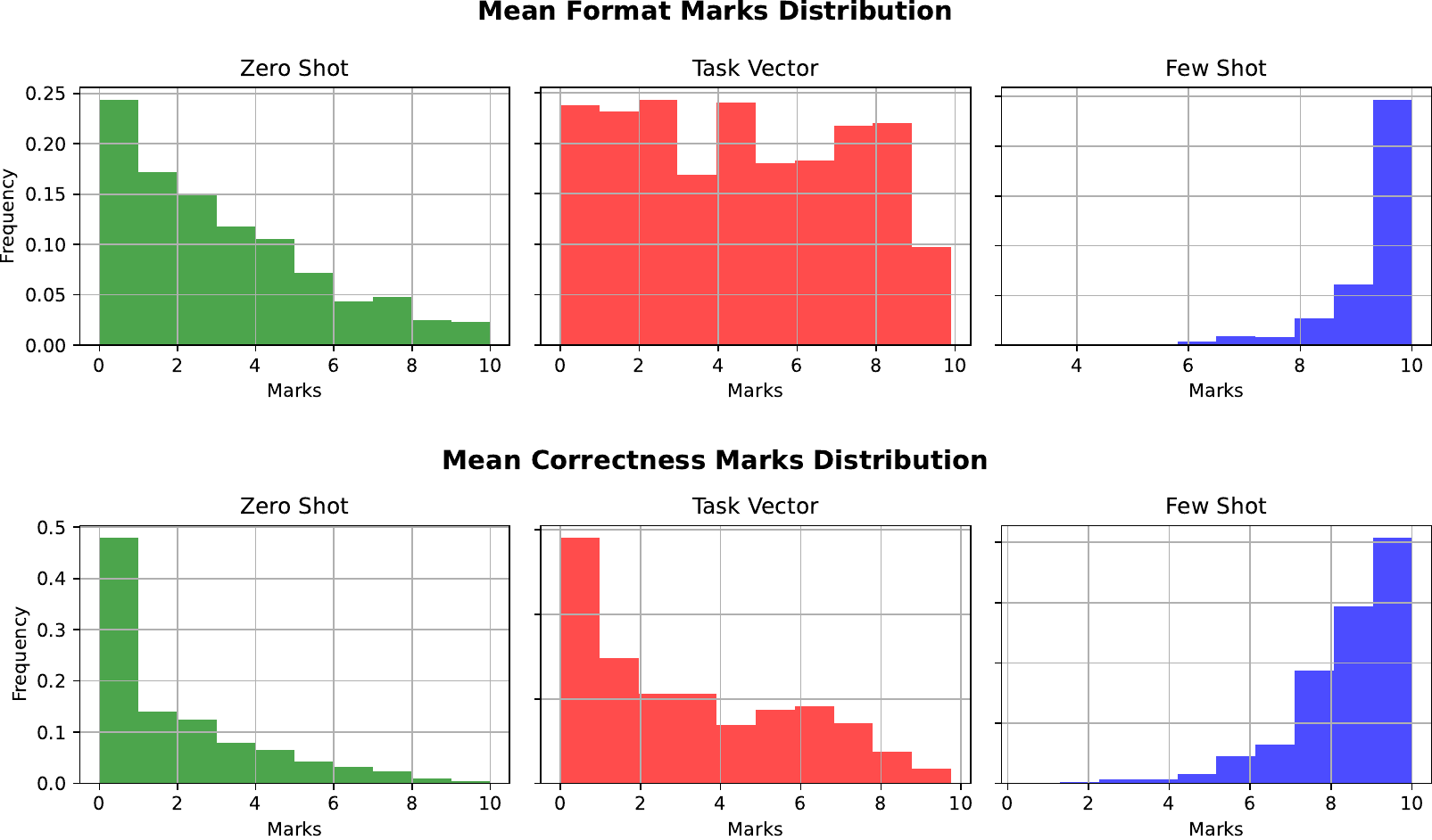}
    \caption{Marks distribution on \qafsc{} dataset.}
    \label{fig:marks_distribution} 
\end{figure*}

However, not all Alpaca entries were suitable for few-shot task generation. Some contained errors (e.g., incorrect calculations or factual inaccuracies), while others were too restrictive (e.g., tasks with limited input diversity). Examples of such problematic entries are listed in Tab.~\ref{tab:alpaca_issues_examples}. To ensure quality, we applied a filtering process to select instructions appropriate for creating diverse, high-quality few-shot examples.

We used Qwen-2.5-72B with a tailored classification prompt (see Appendix~\ref{app:instruction_dataset_classification_prompt}) to evaluate each Alpaca entry's suitability. The prompt assessed whether an \texttt{instruction} and its example \texttt{input} could support generating at least 30 distinct input-output pairs. The evaluation criteria were:
\begin{itemize}
    \item The instruction must allow for $\geq 30$ meaningfully different inputs.
    \item The output's correctness for a given input must be clearly verifiable.
\end{itemize}
Instructions were classified as:
\begin{itemize}
    \item \textbf{GOOD}: Capable of yielding 30+ diverse input-output pairs.
    \item \textbf{LIMITED}: Unsuitable due to insufficient input variety ($<30$).
    \item \textbf{INVALID}: Unsuitable due to reliance on external knowledge, impossibility, or single-output constraints.
\end{itemize}
This process identified 3,096 \textbf{GOOD} instructions for inclusion in \qafsc{}.

For each \textbf{GOOD} instruction, we generated 30-50 new input-output pairs using Qwen-2.5-72B and Qwen-3-235B-A22B~\citep{yang2025qwen3} with a dedicated prompt (see Appendix~\ref{app:few_shot_creation_prompt}). Specifically, Qwen-3-235B-A22B generated 2,072 tasks, and Qwen-2.5-72B generated 1,024 tasks. The prompt instructed the model to analyze the original \texttt{instruction}, \texttt{example\_input}, and \texttt{example\_output} and produce 30 diverse inputs while maintaining the output format and style. The original example served as a template to ensure consistency.

The resulting \qafsc{} dataset comprises 3,096 tasks, each with an original instruction and 30 unique input-output pairs. This structure supports robust few-shot ICL prompts and enables comprehensive analysis of task vector dynamics across diverse task types.

\begin{figure*}[h]
    \centering
    \includegraphics[width=1\textwidth]{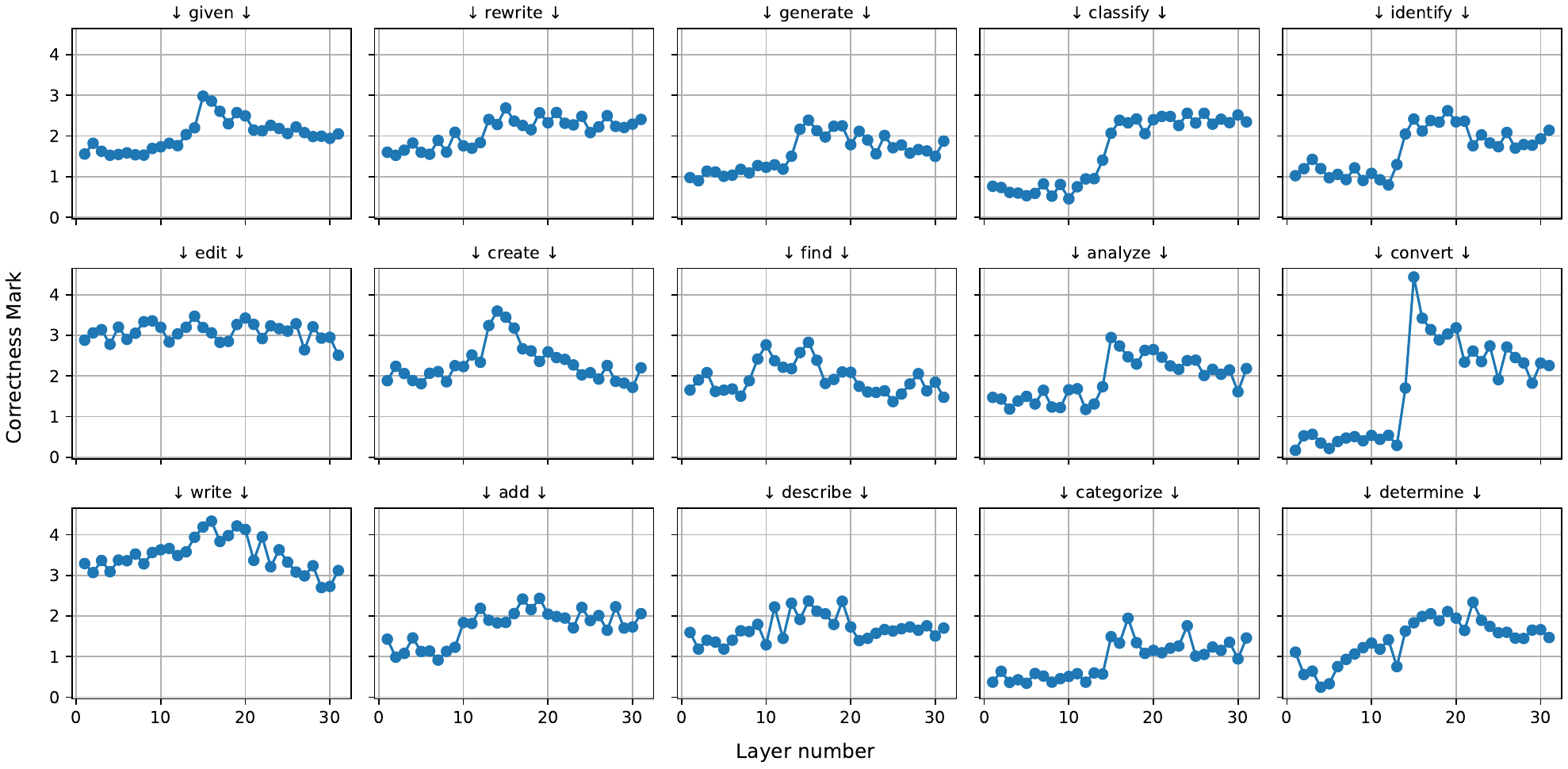}
    \caption{Layer-wise Task Vector Performance across different task categories.}
    \label{fig:category_performance} 
\end{figure*}

\subsection{Task Performance Analysis}
\label{sec:task_performance_analysis}

Prior work evaluating task vectors used tasks with clear, verifiable answers (e.g., antonym generation), enabling simple accuracy metrics. In contrast, \qafsc{} includes diverse tasks, many lacking a single correct output, e.g., rewriting text in a specific style. To uniformly evaluate all tasks, we used an LLM-based judging approach (see Appendix~\ref{app:system_prompt_answer_quality_evaluation}), scoring responses on \textit{format} (0--10, adherence to expected output type) and \textit{correctness} (0--10, accuracy or appropriateness).

For each task, we randomly select 8 examples. Of these, 7 are used to construct the few-shot prompt, while the remaining one serves as the test example for zero-shot evaluation with the task vector. This process is repeated 10 times and the results are then averaged.

We now turn to evaluating how effectively task vectors encode and apply task-specific information across diverse task types. The following experiments investigate task vector performance, layer-wise dynamics, and their limitations in handling complex, multi-faceted tasks.

\section{Experiments and Evaluation}
3,096t learning, we conduct a series of experiments aiming to: (1) assess the layer-wise performance of task vectors across diverse task categories (Sec. \ref{sec:layer-wise_performance}), (2) evaluate their effectiveness compared to zero-shot and full few-shot baselines (Sec. \ref{sec:what_tasks_are_best}).

Unless otherwise specified, all experiments utilize the Llama-3-8B model. First, we analyzed task vector performance across all layers of Llama-3-8B using a randomly selected subset of 1,000 tasks from \qafsc{}. The experimental procedure for processing one such scenario was as follows:

\subsection{Layer-wise Performance}
\label{sec:layer-wise_performance}
\begin{enumerate}
    \item For each layer $l$ in the model, extract the final token's hidden state after processing the few-shot prompt, yielding a layer-specific task vector $v_l$.
    \item Inject $v_l$ into the model's hidden state at layer $l$ at the end of the input for a new zero-shot example.
    \item  Generate an output for this new example with an injected task vector.
    \item Evaluate the output using the LLM-based judging prompts (Appendix~\ref{app:system_prompt_answer_quality_evaluation}), obtaining format and correctness scores.
    \item Average the scores across all evaluated tasks for each layer $l$ and compute per-category averages for key task categories (e.g., Classify, Rewrite, Generate).
\end{enumerate}

\begin{table*}[ht]
\centering
\small
\begin{tabularx}{\linewidth}{@{} p{0.18\linewidth} X @{}}
\toprule
\textbf{Method} & \textbf{Output} \\
\midrule
\multicolumn{2}{@{}X@{}}{\textbf{Task:} Translate the given document into French.} \\
\multicolumn{2}{@{}X@{}}{\textbf{Input:} The novel explores the complexities of human relationships in a dystopian future.} \\
Zero-shot & The novel examines the fragility of human connections in a world where technology has replaced personal interactions. \\
Task Vector & Ce livre explore les complexités des relations humaines dans un futur dystopique. \\
Few-shot & Le roman explore les complexités des relations humaines dans un futur dystopique. \\
\midrule
\multicolumn{2}{@{}X@{}}{\textbf{Task:} Classify the following topic as either technology, business, travel, or sports.} \\
\multicolumn{2}{@{}X@{}}{\textbf{Input:} World Cup football results} \\
Zero-shot & A list of the \\
Task Vector & Sports \\
Few-shot & Sports \\
\midrule
\multicolumn{2}{@{}X@{}}{\textbf{Task:} Re-write the given sentence in the imperative form.} \\
\multicolumn{2}{@{}X@{}}{\textbf{Input:} It's crucial to save water.} \\
Zero-shot & Saving water is crucial. \\
Task Vector & Save water. \\
Few-shot & Save water. \\
\bottomrule
\end{tabularx}
\caption{Examples of tasks from Region 1 (see Fig.~\ref{fig:diff_fig}), where task vectors perform well, showing significant improvement over zero-shot and approaching few-shot performance.}
\label{tab:region1_examples}
\end{table*}

\begin{table*}[ht]
\centering
\small
\begin{tabularx}{\linewidth}{@{} p{0.18\linewidth} X @{}}
\toprule
\textbf{Method} & \textbf{Output} \\
\midrule
\multicolumn{2}{@{}X@{}}{\textbf{Task:} Categorize the given items by type.} \\
\multicolumn{2}{@{}X@{}}{\textbf{Input:} Banana, Pen, Shark} \\
Zero-shot & Banana, Shark, Pen \\
Task Vector & Banana, Pen, Shark \newline Pen, Banana, Shark \newline Pen, Shark, Banana \\
Few-shot & Banana: Fruit, Pen: Writing Instrument, Shark: Animal \\
\midrule
\multicolumn{2}{@{}X@{}}{\textbf{Task:} Analyze the given sentence and classify it into the parts of speech.} \\
\multicolumn{2}{@{}X@{}}{\textbf{Input:} We love to travel in the summer.} \\
Zero-shot & We love to travel in the summer. \\
Task Vector & We love to travel in the summer. \\
Few-shot & We (pronoun) love (verb) to (adverb) travel (verb) in (preposition) the (article) summer (noun). \\

\bottomrule
\end{tabularx}
\caption{Examples of tasks from Region 2 (see Fig.~\ref{fig:diff_fig}), where task vectors perform poorly, often no better than zero-shot, despite few-shot success.}
\label{tab:region2_examples}
\end{table*}

\begin{figure}[t!]
    \centering
    \includegraphics[width=0.5\textwidth]{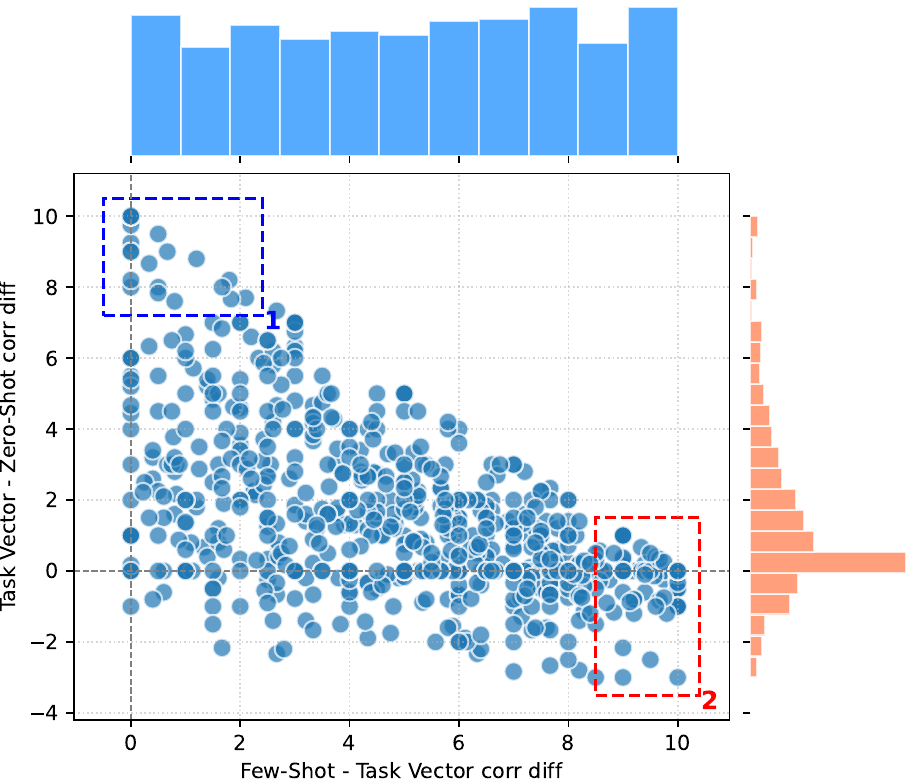}
    \caption{Dual-axis analysis of task vector effectiveness on \qafsc{}, showing performance boost over zero-shot inference versus deficit relative to full few-shot learning.}
    \label{fig:diff_fig} 
\end{figure}

Our analysis, visualized in Fig.~\ref{fig:category_performance}, reveals a consistent performance peak around the 15th layer across different task categories, that correlates with existing results, that intermediate layers are critical for encoding task-specific information. Specifically, task categories such as \textit{convert}, \textit{classify}, \textit{analyze}, etc. exhibit a significant correctness increase at layer 15. However, categories like \textit{edit} and \textit{describe} exhibit more or less the same behavior across all layers.

Since on the 1,000 tasks the most effective was the 15th layer, for all other experiments we stick to only the 15th layer.

The experiments highlight variability in how effectively the model performs in-context learning across the diverse tasks (Fig.~\ref{fig:marks_distribution}). As a sanity check, we compared the performance of task vectors against full few-shot performance and a baseline where no task vector was provided in zero-shot settings.

The Format Score for task vectors consistently exceeded the Correctness Score. This might suggest that the model understood it needed to classify into specific classes (e.g., A, B, C, D) but couldn’t recall what each option represented. Nevertheless, this indicates that the task vector contains some useful signal necessary for task execution, surpassing the baseline, though not as strong as full few-shot performance.

The other thing to notice, that task vectors do not always successfully handle tasks in terms of Correctness Score, which indicates that there is only a small subset of tasks where task vectors perform effectively out of the box. In the following sections we will investigate it more.

\subsection{What tasks are best for task vectors?}
\label{sec:what_tasks_are_best}

Fig.~\ref{fig:diff_fig} illustrates the effectiveness of task vectors by simultaneously measuring their performance boost compared to zero-shot inference and their deficit relative to full few-shot learning. This dual-axis analysis is crucial because strong raw performance from a task vector does not solely indicate its efficacy; it might be an easy task where even zero-shot performs well.

\begin{figure*}[h]
    \centering
    \includegraphics[width=0.9\textwidth]{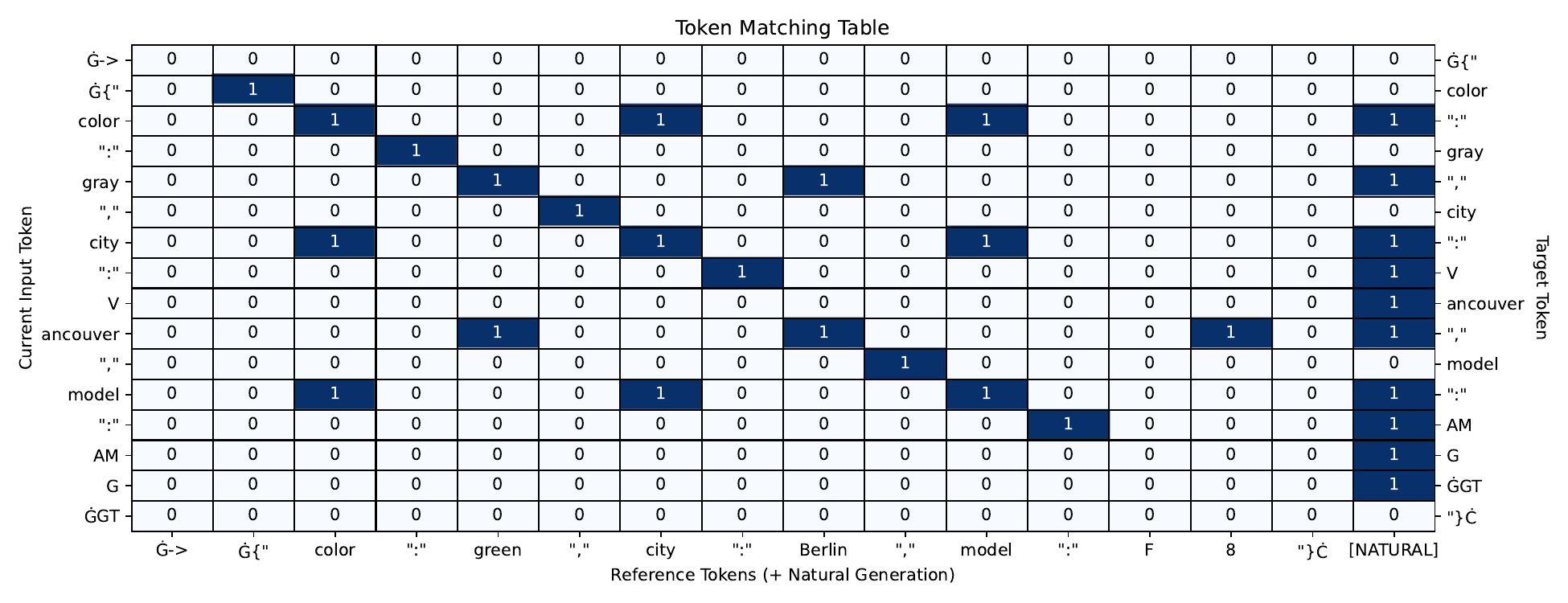}
    \caption{Token-level influence of few-shot hidden states on zero-shot JSON generation. Rows represent tokens being generated in the zero-shot output. Columns represent tokens from the last few-shot example's output, plus a [NATURAL] states for the natural continuation without intervention.}
    \label{fig:json_table} 
\end{figure*}

Tab.~\ref{tab:region1_examples} presents examples illustrative of tasks found in Region 1. In these instances, the application of a task vector leads to outputs that are significantly improved compared to zero-shot and closely match the quality of full few-shot prompting.

Conversely, Tab.~\ref{tab:region2_examples} showcases examples representative of tasks from Region 2. For these tasks, the task vector provides little to no improvement over zero-shot inference, and in some cases, may even lead to a degradation in performance, despite full few-shot prompting demonstrating the task's solvability.

\textit{This motivates a deeper investigation into why task vectors might fail for certain types of tasks.}

\section{Analysis on Complex Tasks}
While task vectors have demonstrated utility, our experiments reveal that their effectiveness is limited. But why do task vectors sometimes fail?
There are at least two possible explanations: (1) all necessary information for task completion is present within the task vector but is obfuscated by noise, or (2) not all the critical task information is captured by the single task vector, instead this information residing elsewhere in the model's representations. 

We hypothesize that many real-world tasks are inherently compound, comprising multiple subtasks. In such cases, a single task vector may not naturally emerge to represent the entire task. Instead, the model develops multiple task vectors—each corresponding to a specific subtask.

\subsection{A Motivating Observation}
To test this hypothesis, we constructed a synthetic dataset that emulates a realistic complex task: converting unstructured textual descriptions into structured schema representations. Specifically, we synthesized automobile descriptions and required the model to transform these descriptions into a predetermined JSON format specified through few-shot examples.

For instance, given an unstructured input description of a car, such as:

\begin{textttbg}
Performance enthusiast's dream: red Rolls-Royce Cullinan EWB (2006). Unleash 564 HP from the 6.75L Twin-Turbo V12, reaching 210 km/h. RWD, Cognac Nappa leather, panoramic roof, rear entertainment, premium audio. Located in Sydney, 3,580 km, VIN: WXZFZXBGE96XAUD55. Priced at \$210,000.
\end{textttbg}

The model was expected to produce a JSON object:
\begin{textttbg}
\{"color":"red","city":"Sydney", "model":"Cullinan"\}
\end{textttbg}

In this setup, the few-shot prompts consist of seven examples, each pairing a unique automobile description with its corresponding JSON representation. These JSON object are always consist of three attributes (each of which is a single token)—\token{color}, \token{city}, and \token{model}—always in that order. 
Here we distinct from the conventional task vector method where a single hidden state intervention occurs after the input description, typically at the \token{->} token. 
Instead, at each step of generating the zero-shot JSON output, we test the influence of injecting \textbf{each} of the hidden states from the output of the last few-shot example. Specifically, for the current token being generated, we make a series of experiments, each time substituting its layer 15 hidden state with \textbf{each} hidden state from \textbf{every token position} within the JSON output of the last few-shot example, and recording the resulting next-token prediction. This is also compared against natural generation without intervention.

To illustrate, Fig.~\ref{fig:json_table} visualizes this process. Rows represent tokens in the output sequence being generated in the zero-shot setting, while columns represent tokens from the output of the last few-shot example. Each cell contains a binary value: \textbf{1} indicates that substituting the hidden state from the column token (from the few-shot example) at layer 15 during generation of the row token (in zero-shot) correctly predicts the next token; \textbf{0} indicates incorrect prediction. This visualization allows us to identify which specific hidden states from the few-shot example contribute to correct predictions for each token in the zero-shot generation.

The token \token{red} in zero-shot was restored when substituting the hidden state of token \token{":"} with the state of the corresponding token \token{":"} in few-shot, where it is followed by token \token{green}.
The token \token{Sy} --- the beginning of the word ``Sydney'' in zero-shot --- was restored from token \token{":"}, followed by token \token{Berlin} in few-shot.
And the token \token{C} --- the beginning of the word ``Culinan'' in zero-shot --- was restored from token \token{":"}, followed by token \token{FB} in few-shot.

The similar behavior holds and for attribute names: the first attribute \token{color} is correctly predicted by substituting the hidden state of the token \token{Ġ\{"}, followed by token \token{color} in few-shot. Same for the second attribute \token{city}, which was restored from the hidden state of the token \token{","}, preceding the token \token{city} from the few-shot example; and for the third:  \token{model}, was restored using token \token{","}, preceding the token \token{model} from the few-shot example.

It is interesting that the first \textit{``real''} task vector was not in the \token{Ġ->}, but in the \token{\{"}.

It is interesting that the first \textit{``real''} task vector was not at the beginning of the output sequence (e.g., associated with the \token{Ġ->} token), but rather emerged at the token \token{\{"}, which directly precedes the first attribute name. This might suggest that the model activates task-specific representations closer to the point of use for each sub-component of the complex output.

\begin{figure}[h]
    \centering
    \includegraphics[width=0.48\textwidth]{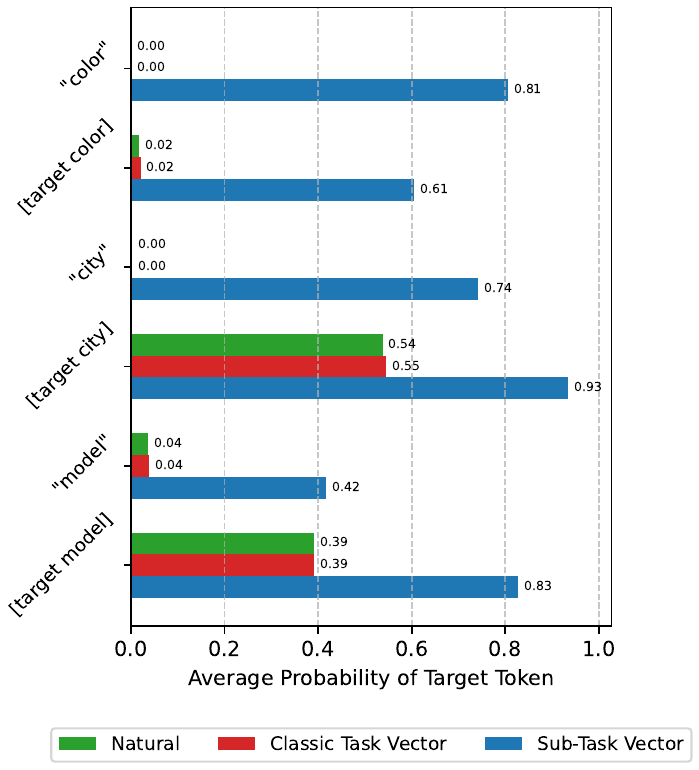}
    \caption{Comparison of token prediction probabilities for JSON generation across 100 automobile descriptions, evaluating Natural Generation, Classic Task Vector, and Sub-Task Vectors strategies.}
    \label{fig:token_probabilities_comparison} 
\end{figure}

\subsection{Investigating Task Vector Specialization}

We further investigate this phenomenon and quantify the potential insufficiency of a single, global task vector for such compositional tasks, we conducted a scaled experiment using a set of 100 (\textit{few-shot, test sample}) pairs of such automobile descriptions. Our evaluation focused on the model's ability to predict specific tokens within the target JSON output. We aimed to measure how effectively different task vector injection strategies could guide the model in generating both \textbf{fixed structural elements} of the JSON, such as tokens for attribute keys (e.g., \token{color}, \token{city}), and \textbf{dynamic, context-dependent content}, namely the attribute values extracted from the input description (e.g., the tokens for color \token{green} or city \token{Berlin}). We assessed performance under three distinct conditions:
\begin{enumerate}
    \item \textbf{Natural Generation}: The model generates the output token by token without any task vector intervention, serving as a baseline.
    \item \textbf{Classic Task Vector}: A single task vector, extracted from the hidden state at the position of the final \token{Ġ->} token in the few-shot prompt's output, is injected at the position of the \token{Ġ->} token in the zero-shot input. This evaluates the conventional single task vector approach.
    \item \textbf{Sub-Task Vectors}: Here we just use the hidden states from the \textit{corresponding} token from the few-shot as a sub-task vectors (it's $i$-th token \token{":"} for attribute values, and tokens \token{Ġ\{"} or \token{","} for attribute names).
\end{enumerate}

The results, summarized in Fig.~\ref{fig:token_probabilities_comparison}. In contrast, and consistent with our motivating observation, the use of sub-task vectors provides a \textbf{substantial} increase in the average probability for the correct target token. This improvement is evident both for predicting fixed attribute \textit{keys} (e.g., token \token{color}, which are identical across examples) and for predicting the dynamic attribute \textit{values} (e.g., the specific color term, which varies between the few-shot demonstration and the zero-shot query). For instance, predicting the token \token{color} after token \token{\{"} sees its probability rise significantly with a sub-task vector, as does the prediction of the actual color value after tokens \token{color} \token{":"}.

It is noteworthy that for later attributes in the sequence, such as the values for \token{city} and \token{model}, the probabilities under Natural Generation are already considerably above random chance. This can be attributed to the model having already processed preceding parts of the JSON structure, thereby gaining contextual cues about the expected format and the current attribute being populated. However, even in these instances where the baseline is stronger, the application of an appropriate sub-task vector still markedly outperforms both Natural Generation and the Classic Task Vector approach. This further reinforces the idea that task execution for complex outputs relies on a sequence of more specialized, context-dependent activations rather than a single, overarching task representation.

\textit{Thus, in this case we need to talk not about \textbf{one} task vector for the entire task, but about \textbf{many} task vectors for the task}.

\section{Conclusion}
\label{sec:conclusion}

To our knowledge, this is a first study to systematically evaluate task vectors on diverse set of NLP tasks.

We found out that optimal task vector performance consistently emerges around a specific intermediate model layer (e.g., the 15th layer of Llama-3-8B) across a wide variety of task types. Second, the overall effectiveness of these vectors varies substantially depending on the intrinsic nature of the task, with some task categories yielding strong performance while others show considerable degradation.

Further, our case analysis of composite tasks reveals a fundamental limitation: a single task vector often fails to capture the full scope of a task. Instead, multiple subtask-specific vectors, distributed across the output sequence, are required to effectively represent and execute complex tasks. This finding challenges the notion that task vectors are inherently noisy approximations of task knowledge, demonstrating that critical task information may be absent from a single vector. 

Future research should therefore explore these distributed and compositional mechanisms of task representation and execution in LLMs to develop a more nuanced understanding of in-context learning.

\section*{Limitations}
First, all experiments were conducted on a single model, Llama-3-8B. Findings regarding optimal layer performance and task-type variability may not generalize to models with different architectures, parameter counts, or from different families.
Second, we employed LLM-based evaluation due to the diversity of tasks in \qafsc{}, precluding the use of a single standard metric like F1-score. This approach, while versatile, can introduce evaluator biases and complicates direct comparisons with studies using task-specific metrics.
Third, our generation process used a fixed temperature for all evaluations. Varying decoding parameters might yield different insights into task vector efficacy, an aspect not explored here.

\section*{Ethics}
This study examines the working mechanisms of large language models and, therefore, does not introduce risks beyond those typically associated with natural language processing or computational linguistics research. 

We utilize the Alpaca dataset \citep{alpaca}, which is licensed under the CC BY-NC 4.0 license, a license suitable for research purposes.

\paragraph{Use of AI Assistants} We utilize Grammarly to enhance and proofread the text of this paper, correcting grammatical, spelling, and stylistic errors, as well as rephrasing sentences. Consequently, certain sections of our publication may be identified as AI-generated, AI-edited, or a combination of human and AI contributions.

\bibliography{references}

\onecolumn

\appendix

\section{Instruction Dataset Classification Prompt}
\label{app:instruction_dataset_classification_prompt}
\begin{lstlisting}[style=promptstyle, caption={Prompt for filtering Alpaca instructions.}, label={lst:filtering_prompt_appendix}]
Your task is to classify each instruction based on how suitable it is for creating few-shot examples. An instruction is good for few-shots if you can generate many different input-output pairs (at least 30) where:
- The same instruction works for all pairs
- Each input is meaningfully different from others
- The output's correctness can be clearly evaluated

Output pure CSV starting with this header:
instruction|example_input|category|explanation

Categories:
GOOD = Good for few-shots: you can create many (30+) valid input examples
LIMITED = Bad for few-shots: cannot generate enough different examples (requires explanation)
INVALID = Invalid: impossible to complete with given input (requires explanation)

Technical rules:
- Start immediately with header
- Process **all** instructions exactly as written, in order
- No quotes in output
- Explain both LIMITED and INVALID cases
\end{lstlisting}

\section{Few-Shot Creation Prompt}
\label{app:few_shot_creation_prompt}
\begin{lstlisting}[style=promptstyle, caption={Prompt for generating few-shot examples.}, label={lst:few_shot_creation_prompt_appendix}]
You are a specialized AI assistant tasked with generating diverse and meaningful examples based on given instructions. Your task is to generate {num_examples} different, high-quality input examples for a given instruction, along with corresponding outputs. Each example should be unique and demonstrate different aspects or applications of the instruction.

Here is the instruction and an example input-output pair for reference:

[INSTRUCTION]
{instruction}

Example format:
Input: {example_input}
Output: {example_output}

Your task is to:
1. Analyze the instruction and understand its scope
2. Generate {num_examples} different, realistic, and diverse inputs that could be used with this instruction
3. For each input, provide an appropriate output following the pattern shown in the example
4. Ensure each input-output pair is unique and demonstrates different aspects of the instruction
5. Format your response exactly as a CSV table with three columns with a header: counter|input|output

Requirements:
- Generate exactly {num_examples} examples
- Ensure all examples are distinct and non-repetitive
- Maintain consistent quality across all examples
- Follow the same style and format as the provided example
- Ensure inputs are realistic and contextually appropriate
- Make outputs match the format and style of the example output

Format your response exactly like this:
```
counter|input|output
1|[first input]|[corresponding output]
2|[second input]|[corresponding output]
...
{num_examples}|[{num_examples}th input]|[corresponding output]
```

Important notes:
- Do not include explanations or additional text
- Start directly with the CSV format
- Use | as separator
- Escape any special characters within the text using double quotes
- Maintain consistent formatting throughout
- Ensure each row follows the exact same pattern
- Do not skip numbers or leave gaps in the counter

Begin your response now by outputting exactly {num_examples} examples in the specified CSV format with | as a separator.
\end{lstlisting}

\section{Answer Quality Evaluation Prompts}
\label{app:system_prompt_answer_quality_evaluation}
\begin{lstlisting}[style=promptstyle, caption={System prompt part of evaluating answer quality prompt.}, label={lst:system_prompt_answer_quality_evaluation}]
You are a judge evaluating responses to tasks. You must provide exactly two scores:

1. Format score (0-10): How well the response matches the semantic type required by the task
- Score 10: Response provides exactly the type of answer requested (e.g., classification label for classification tasks, Yes/No for yes/no questions)
- Score 7-9: Provides the right type of answer with minor formatting issues
- Score 4-6: Partially attempts to provide the required type (e.g., some classifications missing in classification task)
- Score 1-3: Attempts to answer but mostly missing required type markers
.

2. Correctness score (0-10): How accurate/correct the actual answer is within its task domain
- Score 10: Completely correct task completion
- Score 7-9: Mostly correct task completion with minor issues
- Score 4-6: Partial task completion with significant issues
- Score 1-3: Minimal correct task completion
- Score 0: No actual task completion (e.g., just repeating input, missing classifications)

Important: Simply repeating input or providing incomplete answers does not count as task completion. The response must actually perform the requested operation (classify, summarize, etc.) to receive any correctness points.

For classification tasks, any classification label gets a high format score even if wrong (e.g., answering "Opinion" for a fact still gets a high format score). For Yes/No tasks, any Yes/No answer gets a high format score regardless of correctness.

You must output your scores in exactly this format (without quotation marks):
FORMAT_SCORE,CORRECTNESS_SCORE

For example, the output might be:
8,3

Do not provide any explanations or additional text. Only output two numbers separated by a comma.
\end{lstlisting}
\vspace{-0.5cm}
\begin{lstlisting}[style=promptstyle, caption={User prompt part of evaluating answer quality prompt.}, label={lst:user_prompt_answer_quality_evaluation}] 
You are a judge evaluating responses to tasks. You must provide exactly two scores:
1. Instruction: """{instruction}"""
2. Sample Input: """{example_input}"""
3. Expected Output: """{example_output}"""
4. Submission to evaluate: """{output_to_evaluate}"""\end{lstlisting}

\end{document}